\documentclass{llncs}

\usepackage{times}
\usepackage{helvet}
\usepackage{courier}
\usepackage[latin1]{inputenc}
\usepackage[T1]{fontenc}
\usepackage{lscape}

\usepackage[english]{babel}% 
\usepackage[pdftex]{graphicx}
\usepackage{amssymb} 
\usepackage{hyperref}
\hypersetup{colorlinks,%
	citecolor=black,%
	filecolor=black,%
	linkcolor=black,%
	urlcolor=black,%
	pdftex}

\usepackage{times}
\usepackage{epsfig}
\usepackage{float}

\usepackage{booktabs}
\usepackage{array}
\usepackage[ruled,vlined,titlenumbered]{algorithm2e}

\newcolumntype{T}{>{\ttfamily}l} %left type writer

\begin{document}

\newcommand{\pti}[1]{\par\hspace*{0.5em}$ #1 $\hspace*{0.3em}}
\newcommand{\notexists}{\mathchar"7239\!\!\!\raisebox{0.1em}{/\ }}
\newcommand{\point}{\par\hspace*{0.5em}\raisebox{0.2ex}{${\scriptscriptstyle\bullet}$}\hspace*{0.3em}}
\newcommand{\pointlong}{\par\hspace*{1.5em}\raisebox{0.2ex}{${\scriptscriptstyle\bullet}$}\hspace*{0.3em}}
\newcommand{\pointpres}{\par\hspace*{-0.2em}\raisebox{0.2ex}{${\scriptscriptstyle\bullet}$}\hspace*{0.2em}}

\newtheorem{df}{Definition} 
\newtheorem{theo}{Theorem} 
\newtheorem{lem}{Lemma}
\newtheorem{prop}{Proposition} 
\newtheorem{pte}{Property} 
\newtheorem{dompte}{Dominance Property} 
\newtheorem{cor}{Corollary}
\newtheorem{rmq}{Remark}
\newtheorem{notation}{Notation}

\def\fntitrealgo#1{{\normalsize {\sc #1}}}
\def\cstalgo#1{{\sf #1}}
\def\fnalgo#1{{\sc #1}}
\def\fntext#1{{\sc #1}}
\def\sdtext#1{{\it #1}}
\def\sdalgo#1{{\it #1}}
\def\pb#1{{\sf #1}}
\def\mev#1{{\bf{#1}}}
\def\mevdf#1{{\bf{#1}}}
\def\etr#1{{\em #1}}
\def\vrai{{\bf true}}
\def\faux{{\bf false}}
\def\goto{{\bf goto}}
\def\var{{\sl var}}
\def\ind{{\sl ind}}
\SetKw{nil}{nil}
\SetFuncSty{\textit}
%\SetArgSty{\rm}
\SetVline
%\SetKwIf{Ifs}{Elses}{if}{then}{else}{}
%\SetKwFor{Fors}{for}{do}{}
%\SetKwFor{Whiles}{while}{do}{}
%\SetKwArg{Ret}{return}
%\SetKwFor{WhileNot}{tant que non}{faire}{fin tq}
\SetKwFor{Foreach}{for each}{do}{endfor}
%\SetKwIf{If}{Else}{if}{then}{else}{end if}
%\SetKwFor{For}{for}{do}{end for}
%\SetKwFor{While}{while}{do}{end while}
%\SetKwRepeat{Do}{do}{while}
%\SetKwSwitch{Switch}{Case}{Other}{switch}{do}{case}{other}{end switch}
%\SetKwFor{Fn}{}{}{}
\dontprintsemicolon
\SetKwIf{Debut}{EDebut}{}{}{}{}

%
%\title{Two simple and efficient algorithms for maintaining arc consistency of Table and  MDD based constraints}
%\title{The good old discrete relaxation revisited for Table and  MDD based constraints}
\title{Relations between MDDs and Tuples and Dynamic Modifications of MDDs based constraints}
%\title{Improving GAC-4 for maintaining arc consistency of Table and  MDD based constraints}
\author{Guillaume Perez and Jean-Charles R{\'e}gin}
\institute{Universit{\'e} Nice-Sophia Antipolis, CNRS, I3S UMR 7271, 06900 Sophia Antipolis, France\\
  \email{guillaume.perez06@gmail.com, jcregin@gmail.com}}
\maketitle

%%%%%%%%%%%%%%%%%%%%%%%%%%%%%%%%%%%%%%%%%%%%%%%

%\usepackage[latin1]{inputenc}
%\usepackage{graphicx}
%\usepackage[ruled,vlined,linesnumbered]{algorithm2e}

%%%%%%%%%%%%%%%%%%%%%%%%%%%%%%%%%%%%%%%%%%%%
\begin{abstract}
We study the relations between Multi-valued Decision Diagrams (MDD) and tuples (i.e. elements of the Cartesian Product of variables). First, we improve the existing methods for transforming a set of tuples, Global Cut Seeds, sequences of tuples into MDDs. Then, we present some in-place algorithms for adding and deleting tuples from an MDD. Next, we consider an MDD constraint which is modified during the search by deleting some tuples. We give an algorithm which adapts MDD-4R to these dynamic and persistent modifications. Some experiments show that MDD constraints are competitive with Table constraints.
 \end{abstract}
 %%%%%%%%%%%%%%%%

\section{Introduction}

Table constraints are fundamental and implemented in any CP solver. They are explicitly defined by the set of elements of the Cartesian Product of the variables, also called tuples, that are allowed.

Cheng and Yap have proposed to compress the tuple set of the constraint by using Multi-valued Decision Diagrams (MDD) and designed \texttt{mddc} one of the first filtering algorithm establishing arc consistency for them \cite{cheng08,cheng10}. Recently, Perez and Régin have presented MDD-4R a new algorithm which improves \texttt{mddc} \cite{perez14}. MDD-4R proceeds like GAC-4R  and, unlike \texttt{mddc}, maintains the MDD during the search for a solution. They have also introduced an efficient algorithm for reducing a MDD and some powerful algorithms for combining MDDs \cite{perez15}. Thanks to these new algorithms , some experiments based on real life applications shown that the MDD approach becomes competitive with ad-hoc approaches like the filtering algorithms associated with the regular or the knapsack constraints. Thus, the replacement of Table constraints by MDD based constraints appears to be a possible future. In this paper, we wish to take a further step forward in that direction.

We propose to improve the MDD capabilities in three ways. First, we define new methods for transforming a set of tuples, Global Cut Seeds (GCS), sequences of tuples into MDDs. Then, we present some algorithms for adding and removing tuples from an MDD. Next, we consider an MDD constraint which is modified during the search by deleting some tuples.
 
Table constraints are useful for modeling and solving many real-world problems. They can be specified either directly, by input from the user, or indirectly by synthesizing other constraints or subproblems \cite{lhomme12,hoda10}. They have been reinforced in order to deal either from tuples or from sequences of tuples  \cite{focacci01,gent07,regin11b}. Thus, their expressiveness is strong. So, if we want to be competitive with Table constraints we need to be able to represent efficiently different kind of compressed tuple sets. Hence, in the first part of this paper we will show how GCSs and tuple sequences can be represented by an MDD. Notably, we will show that an MDD can be built directly from a sorted list of tuples and that it is not necessary to use an intermediate data structure having a greater space complexity as proposed by Cheng and Yap. In addition, we will show that the MDD representing several disjoint tuple sequences has  a space complexity which is never greater than the one used for expressing the tuple sequences.

Then, we consider the addition and the deletion of one tuple from an MDD. We will see that these operations can be efficiently done by using the method which consists of isolating the path of the MDD corresponding to the tuple in case of deletion and to the common prefix of the tuple in case of addition.
These operations make easier the addition/deletion of a set of tuples, that is the modifications that we can make on an MDD. On one hand, this reinforces the expressiveness of MDDs. On the other hand it also opens the door to dynamic algorithms for maintaining arc consistency of MDDs based constraints. Therefore, we propose an arc consistency algorithm for these constraints when some tuples are definitely deleted during the search for solutions.  
This algorithm is based on previous operations but must be carefully implemented because the deletion of a tuple may create new nodes in the MDD and it causes some problems with the restoration arising after backtracking. In other words, a persistent deletion is a monotonic modification according to the consistency of the constraints, but the maintenance of the MDD is no longer monotonic. 

 Being able to maintain the arc consistency of an MDD whereas some tuples are permanently has two main advantages. First, it is useful for dealing with some problems like nogoods recording. Currently, either ad-hoc algorithms or dynamic table constraint are used. So, it supports our idea to see the MDD constraint as a possible replacement of Table constraints. Second, MDD contraints are now competitive with ad-hoc algorithms of regular constraint. Thus, having a partially dynamic arc consistency algorithms for them gives us immediately a partially dynamic arc consistency algorithm for regular constraints.

The paper is organized as follows. First we  recall background information. Then, we describe how an MDD can be build from tuple sets. Next, we study the addition and the deletion of tuples from an MDD and we present an algorithm for maintaining arc consistency for MDD based constraint when some tuples are permanently deleted during the search for solutions.  
We discuss experiments that empirically establish that using MDD constraints is a competitive approach with using Table constraints. Finally, we conclude this paper.

\section{Background}

Multi-valued decision diagram (MDD) is a method for representing discrete functions. It is a multiple-valued extension of BDDs \cite{bryant86}.
An MDD, as used in CP \cite{andersen07,hadzic08,hoda10,bergman11,gange11},  is a rooted directed acyclic graph (DAG) used to represent some multi-valued function $f : \{0...d-1\}^r \rightarrow \{true,false\}$, based on a given integer $d$ (See Figure 1.). Given the $r$ input variables,
 the DAG representation is designed  to contain $r$ layers of nodes, such that each variable is represented at a specific layer of the graph.  Each node on a given layer has at most $d$ outgoing arcs to nodes in the  next layer of the graph. Each arc is labeled by its corresponding integer. The final layer is represented by the true terminal node (the false terminal node is typically omitted).
There is an equivalence between $f(v_1,...,v_r)=true$ and the existence of a path from the root node to the true terminal node whose arcs are labeled $v_1,...,v_r$. Nodes without any outgoing  arc or without any incoming arc are removed.  

In an MDD constraint, the MDD models the set of tuples satisfying the constraint, such that every path from the root to the true terminal node corresponds to an allowed tuple. Each variable of the MDD corresponds to a variable of the constraint. An arc associated with an MDD variable corresponds to a value of the corresponding variable of the constraint. 

For convenience, we will denote by $d$ the maximum number of values in the domain of a variable; and an arc from $x$ to $y$ labeled by $v$ will be denoted by $(x,v,y)$.

\begin{figure}
	\centering \includegraphics[width=2.5cm]{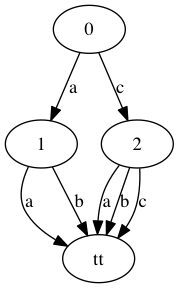}
	\caption{An MDD representing the tuple set \{\{a,a\},\{a,b\},\{c,a\},\{c,b\},\{c,c\}\}}\label{mddFigEx}
\end{figure}

An example of MDD is given in Figure \ref{mddFigEx}. This MDD represents the tuples \{a,a\}, \{a,b\}, \{c,a\}, \{c,b\} and \{c,c\}. For each tuple, there is a path
from the root node (node $0$) to the terminal node (node $tt$) whose are labeled by the tuple values.

The reduction of an MDD is one of the most important operations. It consists of merging equivalent nodes, i.e. nodes having the same set of outgoing neighbors associated with the same labels. 
Usually, a reduction algorithm merge nodes until there is no more any equivalent nodes. 

Most of the time, only reduced MDDs are considered mainly because they are smaller.
Figure \ref{raj} exhibits an MDD having two equivalent nodes: $b$ and $e$. These nodes will be merged by the reduction operation. Note that the reduction operation cannot increase the number of nodes or arcs.
% JE NE VOIS PLUS L'INTERET DE CELA
%, so we have:
%\begin{pte}\label{noIncreaseReduction}
%An MDD $G_2=(X_2,E_2)$ resulting from the reduction of an MDD $G_1=(X_1,E_1)$ satisfies $|X_2| \leq |X_1|$ and $|E_2| \leq |E_1|$.
%\end{pte}

\section{Transformations}

Cheng and Yap have proposed an algorithm for building an MDD from a Table. It uses an internal data structure similar as a trie which requires to have $d$ entries per nodes in order to be able to add a new tuples to the MDD. Then they merge the leaves of the trie and apply the reduction operator. The space complexity is not really an issue with their algorithm because their reduction algorithm also requires to have a direct access to the children of a node, so $d$ entries per node. However, a new reduction algorithm has been recently proposed by Perez and Régin \cite{perez15}. Its space and time complexity are linear, so we can improve the transformation of tuple sets into MDD if we are able to add tuples into an MDD in linear time. In this section, we propose such algorithms.

\subsection{From Trie to MDD}
\begin{figure}
	\centering \includegraphics[width=6.5cm]{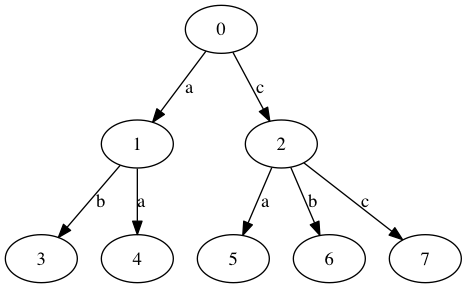}
	\caption{A trie representing the same tuples as the MDD given in Figure \ref{mddFigEx}}\label{trieEx}
\end{figure}

A Trie is a data structure used by Gent et al. for compressing tuple sets \cite{gent07}. Each path from the root to a leaf represents an allowed tuple.
A trie representing a set of $T$ tuples will have $|T|$ leaves. Each variable corresponds to a layer of the trie. A node has a maximum of $d$ children, where $d$ is the size of the domain of the corresponding variable of the node. A example of trie is given in Figure \ref{trieEx}. It corresponds to the same tuple set as the MDD of Figure \ref{mddFigEx}.

A trie can be transformed into an MDD by merging all the leaves into the terminal node $tt$ and by applying the reduction operation \cite{cheng10}.

\subsection{From Table to MDD}

A Table is a data structure where each row represent a tuple and where each column corresponds to a value of a tuple.

Cheng and Yap build and MDD from a Table by defining a trie. Tuples are successively added to the trie. These additions are made by creating first a common node: the root of the trie and by creating paths starting from the root . The rooted subpaths common to several tuples are merged together in order to be represented only once.  Afterwards, all the leaves are merged and the MDD is reduced. 
The drawback of this approach is the addition of a tuple. In order to determine in linear time where it should be added into the existing MDD, the algorithm requires to have $d$ entries per node\footnote{Note that if we accept to increase the time complexity then we can avoid that space complexity.}. 

We propose a simple linear method: we sort the Table and we build the trie from the sorted Table. This can be done efficiently because all tuples are consecutives and so there is no need to search for any position for a tuple: the last one is always the correct one. So we do not need the random access to children and this step can be achieved in linear time. Since the merge of the leaves and the reduction can be performed in linear time too we obtain a linear time algorithm.

\subsection{From GCS and Tuple Sequence to MDD.}
  
A GCS (Global Cut Seed), is a compact representation of a tuple set \cite{focacci01}. A GCS is defined by a set of set of values.  The Cartesian Product of these sets define the represented tuples. It is usually defined as  $c=\{\{v_{1,1},v_{1,2},...,v_{1,k_{1}}\},...,\{v_{n,1},v_{n,2},...,v_{n,k_{n}}\}\}$, where each set of values corresponds to a variable. For instance, given $d=$\{1,2,3,4\}, the GCS $c=\{d,d,d,d\}$ represents the tuple set \{ \{1,1,1,1\}, \{1,1,1,2\},..., \{4,4,4,3\}, \{4,4,4,4\}\}. 
%\begin{center}
%\begin{tabular}{|c|c|c|c|}
%\cline{1-4} 1 & 1 & 1 & 1 \\
%\cline{1-4} 1 & 1 & 1 & 2 \\
%\cline{1-4} ... & ... & ... & ... \\
%\cline{1-4} 4 & 4 & 4 & 4 \\
%\cline{1-4}
%\end{tabular}
%\end{center}
One GCS may represent an exponential number of tuples. However all the tuples cannot be compressed by only one GCS. Two tuples can be represented by the same GCS if they have an Hamming distance equals to 1. For instance, the tuples \{1,1,1\} and \{1,1,2\} may be compressed into  \{1,1,\{1,2\}\}. By contrast the tuples \{1,1,1\} and \{1,2,2\} have an Hamming distance equals to 2 and so cannot be represented by only one GCS. So, the compression of a Table by a set of GCS may required a huge number of GCSs. In order to remedy to this problem, tuple sequences have been introduced \cite{regin11b}.

Tuple sequences generalize GCSs. A tuple sequence encapsulates a GCS and two tuples: a minimum tuple denoted by $t_{min}$ and a maximum tuple denoted by $t_{max}$. It bounds the enumeration of the tuples of the GCS by these two tuples.For instance, let $d$ be the value set \{1,2,3,4\} then the tuple sequence defined by the triplet $s= \{\{d,d,d,d\},\{1,2,2,2\},\{3,1,4,2\}\}$ represents the tuple set
\{\{1,2,2,2\}, \{1,2,2,3\}, ..., \{3,1,4,1\}, \{3,1,4,2\}\}.
%\begin{center}
%\begin{tabular}{|c|c|c|c|}
%\cline{1-4} 1 & 2 & 2 & 2 \\
%\cline{1-4} 1 & 2 & 2 & 3 \\
%\cline{1-4} 1 & 2 & 2 & 4 \\
%\cline{1-4} 1 & 2 & 3 & 1 \\
%\cline{1-4} ... & ... & ... & ... \\
%\cline{1-4} 3 & 1 & 4 & 1 \\
%\cline{1-4} 3 & 1 & 4 & 2 \\
%\cline{1-4}
%\end{tabular}
%\end{center}

Since a tuple sequence is a generalization of a GCS, a method transforming a tuple sequence into an MDD could also be used for transforming a GCS into an MDD.

First, we propose an algorithm for representing one tuple sequence by an MDD. Then, we will show how we can deal with several tuple sequences.
Let $s=(g,t_{min},t_{max})$ be a tuple sequence. For transforming $s$ into an MDD we introduce special nodes: wild card nodes. There is a maximum of one wild card node per layer which is denoted by $w[i]$ for the layer $i$. The wild card nodes are linked together. All the arcs outgoing from $w[i]$ are incoming arcs of node $w[i+1]$ and all arcs outgoing $w_{n-1}$ are incoming arcs of $tt$.

The creation of the MDD representing $s$ is performed in three steps:
\begin{enumerate}
\item The paths corresponding to tuples $t_{min}$ and $t_{max}$ are created. 
\item Arcs from the nodes of the paths previously created to wild card nodes are created as follows. Consider the path created for $t_{min}$. For each layer $i$, let $x[i]$ be the variable corresponding at this layer. Let $val[i]$ be the value set of $g$ for the layer $i$. For each value $a \in val[i]$  such that $a > t_{min}[i]$ we create an arc from the node $n_i$ of the path representing $t_{min}$ to the wild card node $w[i+1]$. We repeat this process for the  path created for $t_{max}$. In addition, we add a particular treatment when a node is shared by the two initial paths, for instance for the root node. In this case, instead of considering all values of $val[i]$, we consider only the values in the interval $val[i] \cap ]t_{min}[i],t_{max}[i][$.
\item From nodes $w[i]$ to node $w[i+1]$ we add as many arcs as there are values in $val[i+1]$.
\end{enumerate}

\begin{figure}
	\begin{center}
\mbox{\epsfxsize=2cm\epsfbox{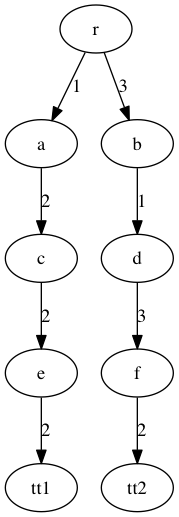}}
\mbox{\epsfxsize=4.5cm\epsfbox{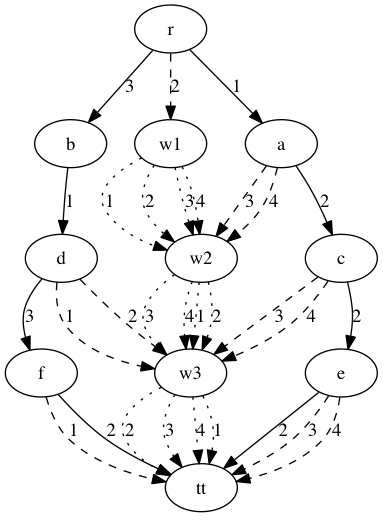}}
\end{center}
	\caption{Transformation of the tuple sequence $s= \{\{d,d,d,d\},\{1,2,2,2\},\{3,1,4,2\}\}$ into an MDD}\label{seqMDD} 
\end{figure}

Figure  \ref{seqMDD} shows the construction of an MDD for the tuple  $s=$ \{\{d,d,d,d\}, \{1,2,2,2\}, \{3,1,4,2\}\} with $d=$\{1,2,3,4\}. The left graph contains the two paths representing the minimum and maximum tuples. The right graph shows how arcs are added to wild card nodes. These arcs are represented by dashed lines. For instance, for node $a$ each value in \{1,2,3,4\} greater than $2$ labels an arc to node $w_1$.
Arcs joining wild card nodes together and with $tt$ are represented by dotted lines. 

Let $r$ be the the number of involved variables. The number of nodes of the obtained MDD is bounded by $3(r-1) +2$. There are $2r$ arcs for the paths corresponding to $t_{min}$ and $t_{max}$. 
There are at most $|val[i]|$ arcs from nodes of the $t_{min}$ path to wild card nodes. We have the same number for the $t_{max}$ path. There are $|val[i+1]|$ arcs from node $w[i]$ to node $w[i+1]$.
Thus, there are at most $\sum_{i=1}^{r}{|val[i]|} + 2r$ arcs in the MDD. This is equivalent to the number of values of the tuple sequence. 

Now, suppose that we have a set of tuple sequences. 
We can consider successively each tuple sequence and build for each sequence MDD with the previous algorithm.
Then, there are two possibilities. Either the tuple sequences are disjoint or not. The former case arises frequently. We just have to merge the graph representation of the MDDs. This can be easily done because they are disjoint. The resulting MDD has a space complexity equivalent to set of tuple sequences and we have:
\begin{pte}
A set of disjoint tuple sequences can be represented by an MDD having an equivalent space complexity.
\end{pte}

The latter case is more complex. A set of disjoint tuple sequences may be computed from a set of non disjoint tuple sequences. Nevertheless, it may create an exponential number of tuple sequences \cite{regin11b}. One possible solution is to partition of the set of tuple sequences into parts of disjoints tuple sequences. Then, each part can be represented by an MDD. If we need to define a constraint then we define a disjunction of MDD constraints. However, the representation of a set of disjoint tuple sequences by a unique MDD without changing the space complexity remains an open question. 

%
% Section ajout suppression
%
\section{Addition and Deletions of tuples from an MDD}

Some work have been carried out for performing operations on BDDs. For instance, Bryant define some algorithms for applying different operators \cite{bryant86,bryant92}. However, the described algorithms are not in-place (i.e. there is the creation of a resulting BDD) and it is not easy to generalize some algorithms designed for BDDs to MDDs mainly because some Booleans rules are no longer true when we have $d$ values in the domain and because the complexity of some algorithms is multiplied by $O(d)$ when dealing with $d$ values. 
Some algorithms have also been proposed for applying operators on MDDs \cite{bergman14,perez15}. However, there are not in-place.

In this section we define in-place algorithms for the addition/deletion of tuples from an MDD. 
Such algorithms are necessary to be able to define a dynamic algorithm for MDD constraint in order to be competitive with dynamic Table constraints.
 
\subsection{Deletion of tuples from an MDD}
First we give an algorithm for deleting a tuple from an MDD. Then, we generalize it.

The deletion of one tuple from an MDD is performed by an operation named path isolation. The idea of this operation is to build a specific path whose arcs are labeled by the values of the tuple that must be deleted. In addition the arcs equivalent of the ones of the isolated path are deleted from the MDD. 
It is performed in four steps:
\begin{enumerate}
	\item The isolation for the first layer
	\item The isolation for any intermediate layer (neither the first nor the last).
	\item The isolation for the last layer
	\item We call an incremental the reduction operator on the MDD
\end{enumerate}

We detail these steps. Let $\tau$ be the tuple that must be deleted. Let $\tau[i]$ be the value for the variable $x[i]$.
\paragraph{Step 1.}  First we identify $a_1=(r,n_1,\tau[1])$ the arc of the first layer labeled by $\tau[1]$ the first value of the tuple. We create the node $ne_1$, the arc $(r,ne_1,\tau[1])$ and we delete the arc $a_1$. We set $mddNode$ to $n_1$ and $isolatedNode$ to $ne_1$. 
\paragraph{Step 2.}  For each layer $i$ from 2 to $r-1$ we repeat the following operation. We identify $a_i=(mddNode,n_{i+1},\tau[i])$ the outgoing arc from the $mddNode$ labeled with $\tau[i]$. We create the node $ne_{i+1}$ and the arc $(isolatedNode, ne_{i+1},\tau[i])$. For each arc $(mddNode, y, w)$ such that $w \neq \tau[i]$ we create the arc $(isolatedNode,y,w)$. We set $mddNode$ to $n_{i+1}$ and $isolatedNode$ to $ne_{i+1}$. 
\paragraph{Step 3.} For each arc $(mddNode, tt, w)$ such that $w \neq \tau[i]$ we create the arc \\
$(isolatedNode,tt,w)$.
\paragraph{Step 4.} We apply the reduction on the MDD by considering only the path and the neighbors of nodes of the path.\\

\begin{figure}
\begin{center}
\mbox{\epsfxsize=1.1cm\epsfbox{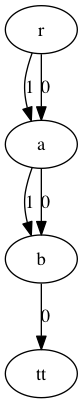}}
\mbox{\epsfxsize=2.7cm\epsfbox{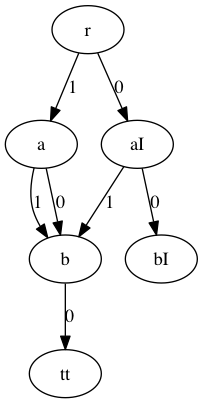}}
\mbox{\epsfxsize=2.4cm\epsfbox{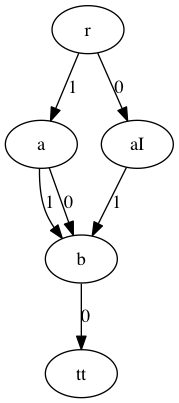}}
\end{center}
\caption{The left MDD is the initial MDD from which we remove the tuple \{0,0,0\}. The middle MDD is the result of the isolation of the path corresponding to the tuple. Nodes $aI$ and $bI$ are created from node $a$ and $b$ in order to create the isolated path. The right MDD is the final result of the operation.}\label{sous1T}
\end{figure}

If at any moment we cannot identify an arc then it means that $\tau$ does not belong to the MDD.
Figure \ref{sous1T} shows the application of this algorithm.

\begin{figure}
\begin{center}
\mbox{\epsfxsize=1.2cm\epsfbox{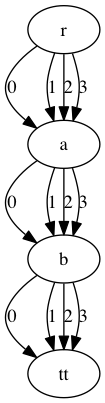}}
\mbox{\epsfxsize=3cm\epsfbox{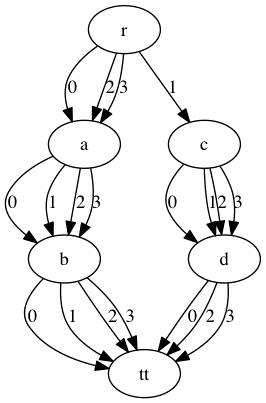}}
\end{center}
\caption{The left MDD represents all the possible tuples for the values  \{0,1,2,3\}. The right MDD represents the deletion of the GCS  \{1,\{0,1,2,3\},1\} from the left MDD. \label{ExSous}}
\end{figure}

The complexity of the deletion of a tuple is bounded by $O(rd)$ because for each isolated node we need to recreate its arcs. However, in practice it is often close to $O(d)$. 
Thus, we can easily implement the deletion of a tuple set by repeating this algorithm. We propose to improve this method

\begin{algorithm}[htb]
\hspace*{-\algomargin}
\fntitrealgo{Deletion}$(L,mdd_1, mdd_2)$ \;
// Step 1: first layer \;
\Foreach{$ (root(mdd_1),v,y_1) \in \omega^{+}(root(mdd_1))$}{
	\If{$\exists (root(mdd_2),v,y_2) \in \omega^{+}(root(mdd_2))$}{
		{\sc addArcAndNode}$(L,1,root(mdd_1),v,y_1,y_2)$ \;
		{\sc deleteArc}$(root(mdd_1),v,y_1)$
	}
}
// Step 2: intermediate layers \;
// $L[i]$ is the set of nodes in layer $i$.\\
%{\sc performOperation}$(L,0,x,mmd1,mdd2,op)$ // pretreatment \;
\Foreach{$i \in 1..r-2$}{
	$L[i] \leftarrow \emptyset$ \;
	\Foreach{node $x \in L[i-1]$}{
		get $x_1$ and $x_2$ from $x=(x_1,x_2)$ \;
		\Foreach{$ (x_1,v,y_1) \in \omega^{+}(x_1)$}{
			\lIf{$\exists (x_2,v,y_2) \in \omega^{+}(x_2)$}{
				{\sc addArcAndNode}$(L,i,x,v,y_1,y_2)$ \;
			}
			\lElse{
				{\sc createArc}$(L,i,x,v,y_1)$ \;
			}
		}
	}
}
// Step 3: last layer \;
\Foreach{node $x \in L[r-1]$}{
	get $x_1$ and $x_2$ from $x=(x_1,x_2)$ \;
	\Foreach{$ (x_1,v,tt) \in \omega^{+}(x_1)$}{
		\lIf{$\not \exists (x_2,v,y_2) \in \omega^{+}(x_2)$}{
			{\sc createArc}$(L,r,x,v,tt)$ \;
		}
	}
}
{\sc pReduce}$(L)$ \;
return $root$ \;
$ $ \;
\hspace*{-\algomargin}
\fntitrealgo{addArcAndNode}$(L,i,x,y_1,v,y_2)$ \;
%\Debut{}{
\If{$\not\exists y \in L[i]$ s.t. $y = (y_1,y_2)$}{
	y $\leftarrow$ {\sc createNode}$(y_1,y_2)$ \;
	add $y$ to $L[i]$ \;
}
{\sc createArc}$(x,v,y)$ \;

\caption{In-place Deletion Algorithm} \label{algoSous}
\end{algorithm}

\subsubsection{Deletion of a set of tuples.} We give an in-place algorithm for deleting a set of tuples from an MDD. In this case, we transform the set of tuples into an MDD and we subtract this new MDD from the initial one. This algorithm generalizes the previous one. It follows the same four steps. It isolates nodes having a common path in both MDD, then it removes the common arcs to the isolated nodes of the second last layer.
Algorithm \ref{algoSous} is a possible implementation.

Figure \ref{ExSous} shows the substraction of the GCS \{1,\{0,1,2,3\},1\} from the  MDD representing all the tuples possibles for the values \{0,1,2,3\}. The GCS is isolated from the MDD. 
Then, the deletion of the arc labeled $1$ of node $d$ correspond to the deletion of only the tuples contained in the GCS.

It is difficult to bound the complexity of the deletion of $T$ tuples, because the MDD created from them may compress the information.

\subsection{Addition of tuples to an MDD}

The addition of tuples into MDD follows the same principles as for the deletion. That is, we also use the idea of path isolation.
We also need to use this idea in order to avoid adding to many things. In other words, we need to precisely control what we add to the MDD.

\begin{figure}
\begin{center}
\mbox{\epsfxsize=3.2cm\epsfbox{SbaseM}}
\mbox{\epsfxsize=4.4cm\epsfbox{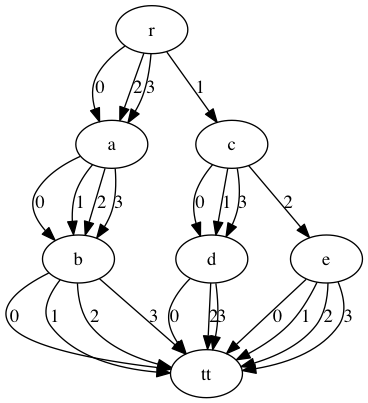}}
\mbox{\epsfxsize=3.5cm\epsfbox{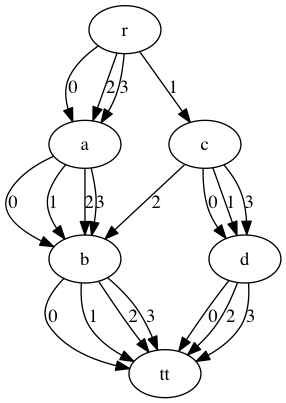}}
\end{center}
\caption{The left MDD is the initial MDD. The middle MDD represents the addition of the tuple \{1,2,1\} to the left MDD, before the reduction. The right MDD is the final MDD obtained after reduction.}\label{raj}
\end{figure}

In this case, the isolated path contains arcs labeled by the values of the tuple that must be added. 
It is performed in four steps:
\begin{enumerate}
	\item The isolation for the first layer
	\item The isolation for any intermediate layer (neither the first nor the last).
	\item The isolation for the last layer
	\item We call an incremental the reduction operator on the MDD
\end{enumerate}

We consider first the addition of one tuple $\tau$. 
The two first steps are very similar as for the deletion. Excepted that at a point, there will be no more path in the MDD having the same subpath as $\tau$. Otherwise, it would mean that $\tau$ is already in the MDD. Thus, at a certain moment we will not be able to identify any arc $(mddNode, n_{i+1}, \tau[i])$ as in step 2 in the deletion algorithm. When this case arises we can stop step 2 and directly create the path from the current isolated node to the terminal node. This path will be labeled by the values of $\tau$ for the remaining layers. Step 3 can be skip. Step 4. remains the same and must be performed.

The complexity of the addition of a tuple is in $O(rd)$ because for each isolated node we need to recreate its arcs.

\begin{algorithm}[htb]
\hspace*{-\algomargin}
\fntitrealgo{Addition}$(L,mdd_1, mdd_2)$ \;
// Step 1: first layer \;
\Foreach{$v \in \omega^{+}(root(mdd_1)) \cup \omega^{+}(root(mdd_2))$}{
	\uIf{$\exists$ $(root(mdd_1),v,y_1)$ $\in$ $\omega^+(root(mdd_1))$ }{
		\If{$\exists$ $(root(mdd_2),v,y_2)$ $\in$ $\omega^+(root(mdd_2))$ }{
			{\sc addArcAndNode}$(L,1,root(mdd_1),v,y_1,y_2)$ \;
			{\sc deleteArc}$(L,i,root(mdd_1),v,y_1)$ \;
		}
	}\lElse{
		{\sc addArcAndNode}$(L,1,root(mdd_1),v,nil,y_2)$ \;
	}
}
// Step 2: intermediate layers \;
// $L[i]$ is the set of nodes in layer $i$.\\
%{\sc performOperation}$(L,0,x,mmd1,mdd2,op)$ // pretreatment \;
\Foreach{$i \in 1..r-2$}{
	$L[i] \leftarrow \emptyset$ \;
	\Foreach{node $x \in L[i-1]$}{
		get $x_1$ and $x_2$ from $x=(x_1,x_2)$ \;
		// If $x_1$ is nil then $\omega^{+}(x_1)$  is empty \;		
		\Foreach{$v \in \omega^{+}(x_1) \cup \omega^{+}(x_2)$}{
			\uIf{$\exists$ $(x_1,v,y_1)$ $\indent$ $\omega^+(x_1)$ }{
				\uIf{$\exists$ $(x_2,v,y_2)$ $\in$ $\omega^+(x_2)$ }{
					{\sc addArcAndNode}$(L,i,x,v,y_1,y_2)$ \;
				} \lElse {
					{\sc createArc}$(L,i,x,v,y_1)$ \;
				}			
			}\lElse{
				{\sc addArcAndNode}$(L,i,x,v,nil,y_2)$ \;
			}
		}
	}
}	
// Step 3: last layer \;
\Foreach{node $x \in L[r-1]$}{
	// If $x_1$ is nil then $\omega^{+}(x_1)$  is empty \;		
	\Foreach{$v \in \omega^{+}(x_1) \cup \omega^{+}(x_2)$}{
		{\sc createArc}$(L,i,x,v,tt)$ \;
	}
}
{\sc pReduce}$(L)$ \;
return $root$ \;
\caption{In-place Addition Algorithm} \label{algoAdd}
\end{algorithm}

\subsubsection{Addition of a set of tuples.} We give an in-place algorithm for adding a set of tuples to an MDD. In this case, we transform the set of tuples into an MDD and we add this new MDD, named $mdd_2$ to the initial one, named $mdd_1$. This algorithm generalizes the previous one. It follows the same steps. Roughly, it isolates nodes having a common path in both MDDs. When a an arc belongs to $mdd_2$ we create a new isolated node and we create an arc from the current isolated node to it. When an arc belongs only to $mdd_1$, we create a arc from the current isolated node to the node in the $mdd_1$.  Algorithm \ref{algoAdd} is a possible implementation.

Figure \ref{raj} shows the effect of the addition of the tuple \{1,2,1\} in the MDD given in Figure \ref{ExSous}. We can see the usefulness of isolated a path for avoiding the addition of the tuples  \{1,\{0,1,3\},1\}. The right MDD shows the impact of the reduction on the MDD: nodes $e$ and $b$ are merged because they have the same outgoing arcs.

It is also difficult to bound the complexity of the addition of $T$ tuples, because the MDD created from them may compress the information.

\paragraph{Duality}: The proximity of these algorithms is due to the duality of the problems: adding a tuple set $T$ to an MDD $M$ is equivalent to delete $T$ from the complementary MDD of $M$.

\subsection{Incremental Reduction}

A reduction step is needed after the deletion and addition of tuples. Using a generic algorithm is costly because it will traverse all the nodes of the MDD and merge the equivalent ones. Since we consider that we add/delete tuples from an MDD which is reduced we can save some computations for the reduction applied after the operation. In fact, it is easy to show that only isolated nodes and nodes having an isolated node as neighbor need to be reconsidered for checking their equivalence. In addition, it is easy to identify isolated nodes because they belongs to the list $L$ of the algorithms. The advantage of this approach is that the reduction step does not increase the complexity of the addition or deletion operations.

%
% fin section ajout suppression
%

%
%MDD4RD
%
\section{Persistent modifications during the search}

Consider $C$ an MDD constraint, we propose to study the problem of the persistent deletions of tuples during the search for solutions and its consequences on the maintenance of the arc consistency of $C$.
In other words we would like to define a kind of dynamic arc consistency algorithm for $C$. It is not a fully dynamic arc consistency algorithm like DnAC-4 \cite{bessiere91} or DnAC-6 \cite{debruyne94} because we will not consider the addition of tuples in the MDD. 

MDD-4R is one of the most efficient algorithm for maintaining the arc consistency of $C$ during the search for solutions.
This is a adaptation of GAC-4 \cite{mohr88b} to MDD which improves the incrementality of GAC-4 \cite{perez14}. For each value $v$ of each variable $x[i]$, MDD-4R maintains $List[i,v]$ the list of the arcs of the layer $i$ labeled by $v$, and belonging to a path from the root to the terminal node. When a value is deleted all the arcs of $List[i,v]$ are deleted from the MDD and these deletions are propagated. That is, each node which has no longer any incoming or outgoing arc is deleted. When a node $n$ is deleted, for instance when it has no more incoming arc, this may lead to the deletion of some arcs, for instance the arcs outgoing from $n$, and MDD-4R checks for each label $v$ whether there is another arc of the same layer having this label. If it is not the case, then the value  $v$ is deleted from the domain of its variable. The strong advantage of MDD-4R is that it does not systematically update the data structures by deleting elements. When it can identify that more than half of the elements of a set will be deleted, it recomputes the set from scratch by adding to it the valid elements. In this case, we say that we ``reset'' the set. This idea can be implemented efficiently with sparse sets and it strongly improves the behavior of the algorithm in practice. 
  
We propose a simple way for dealing with persistent deletions. When a persistent modification arises for a node $n_d$ of the tree search, we will need to also apply this deletion for the ascendant nodes of $n_d$. In addition, we need to be able to backtrack the current MDD from $n_d$ to its parent node and so on... But the deletion of a tuple from an MDD may deeply modified the MDD by adding nodes and removing arcs, so we may have some issues with the backtracking because we will need to reintroduce some arcs. A simple solution for this problem consists of saving the current MDD when a persistent deletion arises for the node $n_d$. Then, when $n_d$ is backtracked, first  we restore the MDD to the saved MDD and we apply the operations required by the backtrack to it. These modifications surely works on the saved MDD. Then, we save the restored MDD and redo all the persistent deletions that have been made from node $n_d$ during the search for a solution.
This methods also works well for Table constraint because deleting a tuple is quite simple.

Note that it is required to restart the arc consistency algorithm associated with each constraint impacted by the persistent deletion after a backtrack because some values may be no longer valid.

This algorithm can be improved but the algorithm becomes quite complex and is out of the scope of this paper. 
%
% fin MDD4RD
%

\section{Experiments}

The goal of these experiments is not to show that MDD-4R is a competitive algorithms compared with GAC-4R or some other efficient algorithms for Table constraints, even if we recall some results. This study has already been done in \cite{perez14}. We aim at showing that even when performing deletions, an MDD approach is competitive with a Table approach, even if it uses compressed tuples.

\paragraph{Machines} MacBook Pro, Intel Core I7, 2,3GHz, 8GB memory. 
\paragraph{Solveur} or-tools 3158.

\paragraph{Selected Instances} We build random instances for having a global pictures of the behaviors of the two approaches. 

\begin{figure}
	\centering \includegraphics[width=6.5cm]{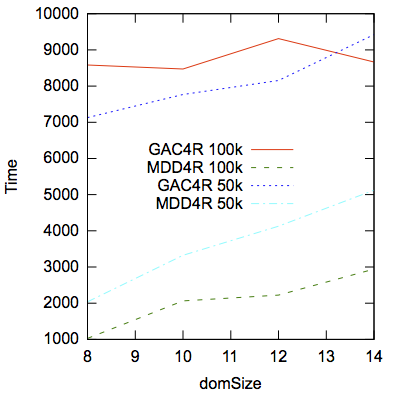}
	\caption{Search for one solution in a problem involving 5 constraints. One involving 8 variables, the other involving 4 variables}\label{res1}
\end{figure}

\paragraph{GAC-4R vs MDD-4R}
We recall the advantage of MDD-4R over GAC-4R when there are a large number of tuples (See Figure \ref{res1}).

\begin{figure}
\begin{center}
\mbox{\epsfxsize=5cm\epsfbox{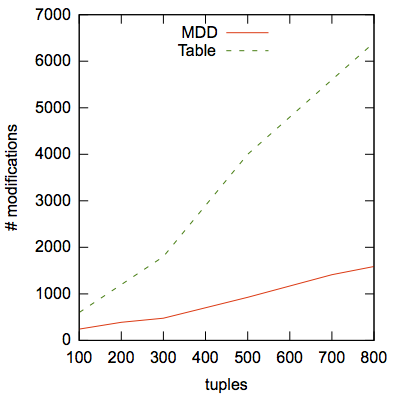}}
\mbox{\epsfxsize=7cm\epsfbox{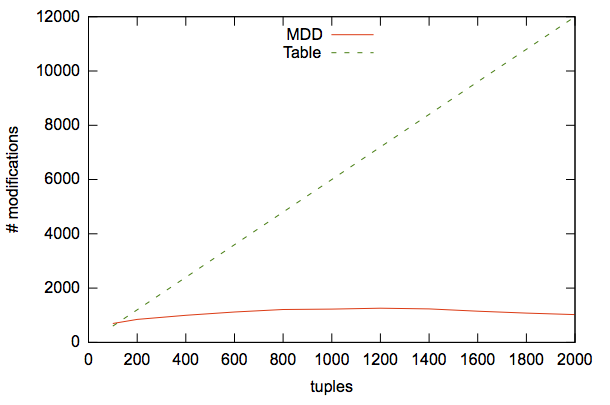}}
\mbox{\epsfxsize=6cm\epsfbox{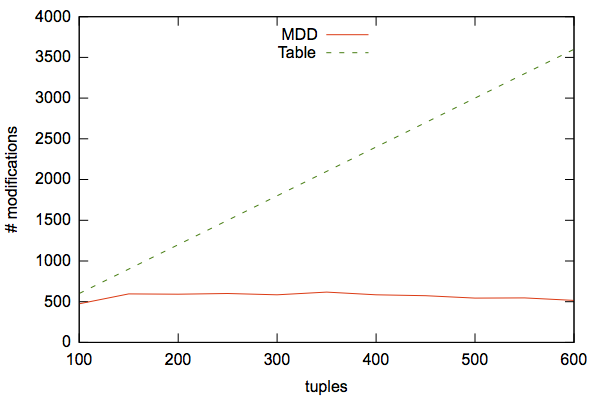}}
\end{center}
\caption{Number of modifications triggered by the deletions of tuples. 
Top left graph: medium density. Top right graph: high density. Bottom graph: high density}\label{res2}
\end{figure}

\paragraph{Impacts of Deletions}

We study the number of modifications triggered by the deletions of tuples. The tuple set  involves 6 variables and 5 values. Figure  \ref{res2} shows the results for three types of tuple sets one having a low tightness (8\%), medium  tightness (15\%) and high  tightness (25\%). We observe a linear evolution for the Table constraint which is normal. The MDD approach requires less modifications. Surprisingly, the best results are obtained for the low or high densities. The worst case for MDDs seems to be a medium density. This can be explained by the fact that the modifications have a lot of work to do and there are a lot of nodes. High density are strongly compressed, whereas there are less work to do for low density.

\paragraph{Deletions of tuples during the search}

We consider a problem involving constraints of arity $6$. Each constraint is defined from a Table having 230 000 tuples. We search for all solutions and perform 100 000 persistent modifications during this search. This lead to 600 000 modifications for the Table constraints, because we need to check whether a deleted tuple was a support or not for each of its value. Interestingly, the number of modifications ( creation/deletion of arcs and nodes) triggered for the MDD is 135 000 . This is smaller. This comes from the fact that an MDD compresses the tuples. Thus, more operations may be required when one tuple is deleted but the data structure remains compressed and so remains more powerful. The following table gives us some information about the time needed for performing these operations. Once again, the MDD approach performs well.

\begin{table}
\begin{center}
\begin{tabular}{|l|r|r|r|r|}

\cline{1-5} Algorithm & \#deletions & \multicolumn{3}{|c|}{Domain size}\\
\cline{1-5} & & 10 & 12 & 14\\

\cline{1-5} GAC4R & 0 & 1570 & 2123 & 2710 \\
\cline{1-5} GAC4R & 100 000 & 1274 & 1511 & 1763 \\
\cline{1-5} MDD4RD & 0 & 1115 & 1385 & 1815 \\
\cline{1-5} MDD4RD & 100 000 & 763 & 879 & 1064 \\

\cline{1-5}
\end{tabular}
\end{center}
\caption{Time (in ms) for deleting 100 000 tuples from tuple set containing 230 000 elements during the search for all solutions. \label{figtab1}}
\end{table}

\section{Conclusion}
We have given an algorithm for transforming tuple sets, GCS and tuple sequences into an MDD. Then, we have described efficient in-place algorithms for adding or deleting tuples from an MDD. At last, we have considered the dynamic modification of an MDD constraint and proposed a dynamic algorithm for maintaining the arc consistency during the search for solution. We have also shown some experiments. 
This works contributes to the proof that an MDD approach is competitive with a Table approach for representing constraints in extension in constraint programming.

%%%%%%%%%%%%%%%%%%%%%%%%%%%%%%%%%%%%%%%%%%%%

%\subsubsection{Acknowledgments.}

% TODO : il faudra remercier Christophe Lecoutre

%We would like to thank very much Laurent Perron for his useful comments which helped to improve the paper

%%%%%%%%%%%%%%%%%%%%%%%%%%%%%%%%%%%%%%%%%%%%

%\pagebreak

\bibliographystyle{plain}

\bibliography{jcr}

\begin{thebibliography}{10}

\bibitem{andersen07}
Henrik~Reif Andersen, Tarik Hadzic, John~N. Hooker, and Peter Tiedemann.
\newblock A constraint store based on multivalued decision diagrams.
\newblock In {\em CP}, pages 118--132, 2007.

\bibitem{bergman14}
D.~Bergman, A.~Cire, and W-J. van Hoeve.
\newblock Mdd propagation for sequence constraints.
\newblock {\em Journal of Artificial Intelligence Research}, 50:697--722, 2014.

\bibitem{bergman11}
David Bergman, Willem~Jan van Hoeve, and John~N. Hooker.
\newblock Manipulating mdd relaxations for combinatorial optimization.
\newblock In {\em CPAIOR}, pages 20--35, 2011.

\bibitem{bessiere91}
C.~Bessi{\`e}re.
\newblock Arc-consistency in dynamic constraint satisfaction problems.
\newblock {\em Proceedings AAAI}, 1991.

\bibitem{bryant92}
R.~E. Bryant.
\newblock Symbolic boolean manipulation with ordered binary decision diagrams.
\newblock {\em ACM Computing Surveys}, 24(3):293--318, 1992.

\bibitem{bryant86}
Randal~E. Bryant.
\newblock Graph-based algorithms for boolean function manipulation.
\newblock {\em IEEE Transactions on Computers}, C35(8):677--691, 1986.

\bibitem{cheng10}
K.~Cheng and R.~Yap.
\newblock An mdd-based generalized arc consistency algorithm for positive and
  negative table constraints and some global constraints.
\newblock {\em Constraints}, 15, 2010.

\bibitem{cheng08}
Kenil C.~K. Cheng and Roland H.~C. Yap.
\newblock Maintaining generalized arc consistency on ad hoc r-ary constraints.
\newblock In {\em CP}, pages 509--523, 2008.

\bibitem{debruyne94}
R.~Debruyne.
\newblock Dn-ac6.
\newblock Master's thesis, University Montpellier II, 1994.

\bibitem{focacci01}
F.~Focacci and M.~Milano.
\newblock Global cut framework for removing symmetries.
\newblock In {\em Proc. CP'01}, pages 77--92, Paphos, Cyprus, 2001.

\bibitem{gange11}
G.~Gange, P.~Stuckey, and Radoslaw Szymanek.
\newblock Mdd propagators with explanation.
\newblock {\em Constraints}, 16:407--429, 2011.

\bibitem{gent07}
I.~Gent, C.~Jefferson, I.~Miguel, and P.~Nightingale.
\newblock Data structures for generalised arc consistency for extensional
  constraints.
\newblock In {\em Proc. AAAI'07}, pages 191--197, Vancouver, Canada, 2007.

\bibitem{hadzic08}
Tarik Hadzic, John~N. Hooker, Barry O'Sullivan, and Peter Tiedemann.
\newblock Approximate compilation of constraints into multivalued decision
  diagrams.
\newblock In {\em CP}, pages 448--462, 2008.

\bibitem{hoda10}
Samid Hoda, Willem~Jan van Hoeve, and John~N. Hooker.
\newblock A systematic approach to mdd-based constraint programming.
\newblock In {\em CP}, pages 266--280, 2010.

\bibitem{lhomme12}
Olivier Lhomme.
\newblock Practical reformulations with table constraints.
\newblock In {\em ECAI}, pages 911--912, 2012.

\bibitem{mohr88b}
R.~Mohr and G.~Masini.
\newblock Good old discrete relaxation.
\newblock In {\em Proceedings of ECAI-88}, pages 651--656, 1988.

\bibitem{perez14}
G.~Perez and J-C. R{\'{e}}gin.
\newblock Improving {GAC-4} for table and {MDD} constraints.
\newblock In {\em Principles and Practice of Constraint Programming - 20th
  International Conference, {CP} 2014, Lyon, France, September 8-12, 2014.
  Proceedings}, pages 606--621, 2014.

\bibitem{perez15}
G.~Perez and J-C. R{\'e}gin.
\newblock Efficient operations on mdds for building constraint programming
  models.
\newblock In {\em International Joint Conference on Artificial Intelligence,
  IJCAI-15}, Argentina, 2015.

\bibitem{regin11b}
J-C. R{\'e}gin.
\newblock Improving the expressiveness of table constraints.
\newblock In {\em CP'11, proceedings workshop ModRef'11}, 2011.

\end{thebibliography}
%\bibliography{jcr}

\end{document}